\documentclass[conference]{IEEEtran}
\IEEEoverridecommandlockouts
\usepackage{amsmath,amssymb,amsfonts}
\usepackage{algorithmic}
\usepackage{graphicx}
\usepackage{textcomp}
\usepackage{hyperref}
\usepackage{xcolor}
\usepackage{subfig}
\usepackage[numbers,sort&compress]{natbib}

\makeatletter
\newcommand{\newlineauthors}{%
  \end{@IEEEauthorhalign}\hfill\mbox{}\par
  \mbox{}\hfill\begin{@IEEEauthorhalign}
}
\makeatother

\def\BibTeX{{\rm B\kern-.05em{\sc i\kern-.025em b}\kern-.08em
    T\kern-.1667em\lower.7ex\hbox{E}\kern-.125emX}}
\begin{document}

\title{RCURRENCY: Live Digital Asset Trading Using a Recurrent Neural Network-based Forecasting System}


\author{\IEEEauthorblockN{Yapeng Jasper Hu}
\IEEEauthorblockA{\textit{Delft University of Technology}\\
\textit{Distributed Systems Group} \\
Delft, The Netherlands \\
y.j.hu@student.tudelft.nl}
\and
\IEEEauthorblockN{Ralph van Gurp}
\IEEEauthorblockA{\textit{Delft University of Technology}\\
\textit{Distributed Systems Group} \\
Delft, The Netherlands \\
r.v.gurp@student.tudelft.nl}
\and
\IEEEauthorblockN{Ashay Somai}
\IEEEauthorblockA{\textit{Delft University of Technology}\\
\textit{Distributed Systems Group} \\
Delft, The Netherlands \\
a.somai@student.tudelft.nl}
\newlineauthors
\IEEEauthorblockN{Hugo Kooijman}
\IEEEauthorblockA{\textit{Delft University of Technology}\\
\textit{Distributed Systems Group} \\
Delft, The Netherlands \\
h.v.kooijman@student.tudelft.nl}
\and
\IEEEauthorblockN{Jan S. Rellermeyer}
\IEEEauthorblockA{\textit{Delft University of Technology}\\
\textit{Distributed Systems Group} \\
Delft, The Netherlands \\
j.s.rellermeyer@tudelft.nl}
}

\maketitle

\begin{abstract}
Consistent alpha generation, i.e., maintaining an edge over the market, underpins the ability of asset traders to reliably generate profits. Technical indicators and trading strategies are commonly used tools to determine when to buy/hold/sell assets, yet these are limited by the fact that they operate on \emph{known} values. Over the past decades, multiple studies have investigated the potential of artificial intelligence in stock trading in conventional markets, with some success. In this paper, we present RCURRENCY, an RNN-based trading engine to predict data in the highly volatile digital asset market which is able to successfully manage an asset portfolio in a live environment. By combining asset value prediction and conventional trading tools, RCURRENCY determines whether to buy, hold or sell digital currencies at a given point in time. Experimental results show that, given the data of an interval \textit{t}, a prediction with an error of less than 0.5\% of the data at the subsequent interval \textit{t+1} can be obtained. Evaluation of the system through backtesting shows that RCURRENCY can be used to successfully not only maintain a stable portfolio of digital assets in a simulated live environment using real historical trading data but even increase the portfolio value over time.
\end{abstract}


\section{Introduction}
Ever since the emergence of modern financial markets in the 17th century, specialists have made an effort to maximize profits from trading goods, services, stocks and securities. Consequently, consistent alpha generation \cite{gross2005consistent}, a term to describe the edge over the market or the ability to make profit, has remained a hot topic \cite{fung2002news, saad1998comparative}. With the advent of the digital age and the ability to process large amounts of data, many strategies have revolved around new techniques and computational breakthroughs in an effort to predict the upward and downward trends of assets (e.g., \cite{lee2009using, ni2011stock}). The reliability of these predictions often remains questionable \cite{fama1995random, vui2013review}. This difficulty in predicting traditional asset data is all the more true for the recent digital asset (e.g. Bitcoin) markets, which have proven to be extraordinarily volatile \cite{ali2014economics, dyhrberg2016bitcoin, yermack2015bitcoin}.

However, the recent rise of AI and ML has opened up new opportunities. AI techniques can analyze enormous amounts of data and observe decades of conventional experience at a rate that is only limited by the complexity of the technique and the available computational power \cite{bottou2010large, bottou2018optimization}. Furthermore, modern machine learning algorithms are capable of discovering patterns in data that are hidden to the human eye due to the volume and complexity of the data \cite{witten2016data}. Unsurprisingly, investment and asset management firms have started to invest billions into developing AI-based asset trading and investment applications \cite{bloomberg2019quants}.

In this paper, we present our findings and experiences in designing and implementing \emph{RCURRENCY}, an automated trading engine tasked with managing a portfolio of digital assets (Section \ref{sec:model}). The primary goal of our research is to provide insights into the capabilities of AI-powered digital asset trading in a highly volatile and therefore challenging live market environment. Effectively managing a portfolio requires the AI-based training engine to estimate when to sell off assets, buy new ones, or hold the current position. Decisions are made based on data gathered from a digital asset exchange at a regular interval. 
For this purpose, we combine the prediction with previous (known) values and apply a trading strategy to derive and execute a final buy/hold/sell decision. In practice, achieving a stable portfolio is considered a high bar for automatic trading engines to master.

We experimentally evaluate different trading strategies based on actual historical market data (Section \ref{sec:experiments}). Our results show that the prediction accuracy is highly dependent on the configuration of the neural network. The Root Mean Squared Error over a prediction can increase significantly when incorrect parameters are chosen. 
We thoroughly evaluate the most important parameter spaces and their impact on the prediction accuracy. When an adequate set of parameters is chosen, the trading engine is capable of estimating future asset values with an error of approximately 0.4\%. In practice, this makes it possible to reliably predict up and down motions, as well as approximate the absolute change in value. This, in turn, allows the system to maintain and increase portfolio value over time. 


\section{Related Work}

The suggestion of predicting stock prices utilizing historic data has been an active research topic for decades. One of the early prediction methods based on an inter-temporal general equilibrium model was proposed by \citet{balvers1990predicting}.  These early methods used a purely analytical approach applying basic mathematics and statistics. However, some schools of thought even then disputed the merit of the approach and results of predicting the stock prices. For instance, it has been empirically shown that if a random walk strategy accurately reflects the reality, then all models concerning (stock) market predictions are meaningless~\cite{fama1995random}.

Nowadays, more variables are added to the prediction model, such as the market volume \cite{brooks1998predicting}, twitter sentiment \cite{zhang2011predicting}, investor sentiment, or media news content \cite{tetlock2007giving}. 
Whereas fundamental analysis describes the examination of data focused on macro-economic variables \cite{abarbanell1997fundamental}, technical analysis puts more emphasis on extracting trends and patterns of an investment instrument's price, volume, breadth and trading activities \cite{thomsett1999mastering,pring1991technical,meyers1994technical}.
The methodology proposed and analyzed in this paper focuses on technical analysis while it could be expanded to support fundamental analysis in future works. Most recently, multiple studies have used AI techniques for prediction. Examples of concrete techniques include particle swarm optimization \cite{nenortaite2004stocks}, nearest neighbor classification \cite{teixeira2010method}, and neural networks \cite{lam2004neural}. An empirical study by \citet{gu2018empirical} compared multiple AI techniques in combination with trading techniques in the conventional stock market. Their fundamental result is that overall AI can be successfully utilized and the predictive capability of the AI techniques carries a significant advantage. However, using AI in highly volatile markets such as digital assets \cite{dyhrberg2016bitcoin} has gone largely unexplored to this date.

Some initial efforts to predict prices and trends of Bitcoin and other digital currencies have been undertaken by \citet{mcnally2018predicting} and \cite{alessandretti2018anticipating}. \citet{mcnally2018predicting} explore prediction of Bitcoin prices based on the Simple Moving Average (SMA) technical indicator, through RNN, LSTM and ARIMA methods. 

\citet{alessandretti2018anticipating} investigate price prediction of a large number of cryptocurrencies using tree boosting techniques, as well as an LSTM-based method. The researchers simulate trading of currencies, taking into account transaction fees, but fail to explore the myriad of strategies traders use to determine their investments. In either case the number of input variables is limited; resulting the loss of potentially valuable network input data.

\section{The RCURRENCY Live Trading System} \label{sec:model}

\begin{figure*}
    \centering
    \includegraphics[width=0.82\textwidth]{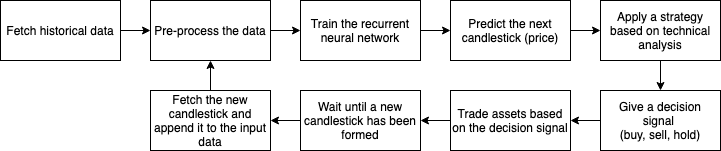}
    \caption{Structure and data flow in RCURRENCY.}
    \label{fig:rcurrency_flow}
\end{figure*}

In order to address the challenge of making reliable predictions of cryptocurrency markets for live trading, we designed the RCURRENCY system and structured it into three separate components. The responsibility of the first component is to collect and preprocess the timely-ordered series of data. The second component feeds this data into the neural network and, by doing so, trains it to recognize patterns in the processed data.
These two components are sufficient to grant the system predictive capabilities for single steps in the time series data. The third component allows the system to make trading decisions depending not only on a single prediction, but also on an elaborate analysis of trends and patterns. It implements and applies various trading strategies, each of which calculates a trading signal based on its respective predefined rules. The flow diagram of the RCURRENCY system is visualized in Figure~\ref{fig:rcurrency_flow}.

\subsection{Training Data}

A neural network-based prediction method is only as reliable as the data that can be leveraged to train the network. As a result, an essential step in the processing pipeline is to collect the raw data and turn it into preprocessed training data for the learning algorithm to digest. Our system obtains raw Bitcoin training data by calling an API provided by CryptoCompare\footnote{https://min-api.cryptocompare.com/}. This company offers services concerning over 3.000 digital assets, including the ability to retrieve historical price data. Such data comprises candlestick data points in one-minute time frames, where each data point contains four values: the \emph{open}, \emph{high}, \emph{low}, and \emph{close} price of the time frame. The system runs a simple script to recursively and sequentially call this API to fetch historical Bitcoin price data, which consists of approximately 80.000 hourly data points, spanning back to August 2010. In contrast, live data on the current market condition is instead fetched directly from a cryptocurrency exchange to minimize acquisition delays. 


The raw data then undergoes preprocessing. Unfortunately, approximately the first 30 percent of the data set contains many near-zero values that tend to deteriorate the prediction capabilities of the network. These values do not contain sufficient variance for the gradient descent of the error function to converge. As a consequence, all data up until May 2013 has been manually removed, roughly up until the time Bitcoin reached values around one hundred USD.

Next, the system applies some technical analysis on the resulting data set, which involves extracting trends and patterns from the price data by constructing technical trading indicators. Each indicator is calculated according to their respective formula, where each formula uses some (or all) of the data set as input. RCURRENCY implements the financial technical indicators.
The calculated indicator values are added as additional rows to the training data set. This extra training data provides the neural network with more information about the market and should thus lead to higher predictive accuracy.


The data is subsequently converted to what is called a \emph{stationary data set}. Such a data set is created by taking the difference of each data point from the previous data point in a column-based manner. 


Finally, row-based normalization is applied to the data set to fit the resulting values in the range of -1 and 1 in order to obtain better results and reduce the training time \cite{sola1997importance}. 
The new row-normalized matrix is mathematically defined as follows:

Given a matrix $x$, with dimensions $m \times n$, for the new matrix $y$ holds:
\begin{equation}
y_{i,j} = 2 * (\dfrac{x_{i,j} - y^{min}}{y^{max} - y^{min}}) - 1, 1 \leq i \leq m, \& 1 \leq j \leq n.
\end{equation}

\subsection{Recurrent Neural Network}

There are various options available regarding networks capable of consuming a series of data points in order to make predictions on future values, such as the time series analysis technique ARIMA \cite{mills1991time}, or neural networks. As evaluated by \citet{kohzadi1996comparison}, a neural network outperforms the ARIMA network over longer periods of time, justifying its preference for our system. Specifically a recurrent neural network functions as RCURRENCY's predictor, which is a particular type of neural network capable of exhibiting temporal dynamic behavior. As first introduced by \cite{mikolov2010recurrent}, these networks model the connections between layer nodes as a directed graph along a temporal sequence. The input data for RCURRENCY's recurrent neural network consists of so-called candlesticks, data structures containing numerical stock market information on assets. The values in a candlestick are dependent on previous candlestick values and do therefore not follow a random walk \cite{lo1988stock}.

A conventional recurrent neural network experiences problems with the gradient descent when learning long-time dependencies, as depicted in the research by \citet{bengio1994learning}. When an LSTM layer is used in the architecture of the recurrent neural network, the problems of exploding and vanishing gradients are solved \cite{Hochreiter1997}. An example of a long-term dependency in the current context is given by \citet{tsinaslanidis2014prediction}. Their paper shows that a previously encountered subsequence of the time series data could be similar to the current trend. The consequent actions of the price following the previously encountered subsequence could carry value in the predictions of the actions following the current subsequence.

The recurrent neural network encompasses three layers:


\begin{enumerate}
    \item \textit{Linear layer}\\
    The linear layer is required to map the nodes of the input layer to the number of nodes in the hidden layer. It is a fully connected layer, of which every edge has an individual weight.
    
    \item \textit{Fast LSTM layer}\\
    The FastLSTM layer is a special, faster version of the conventional LSTM layer. It combines the calculation of the input, forget and output gates and the hidden state in a single step.

The composite functions of the standard LSTM cell are based on the paper from \cite{graves2013speech}.
and are defined as follows:
    \begin{equation}
        \begin{split}
            i_t = \sigma(W_{xi}x_t + W_{hi}h_{t-1} + W_{ci}c_{t-1}+b_i)\\
            f_t = \sigma(W_{xf}x_t + W_{hf}h_{t-1} + W_{cf}c_{t-1}+b_f)\\
            c_t = f_tc_{t-1} + i_t\text{tanh}(W_{xc}x_t + W_{hc}h_{t-1} + b_c)\\
            o_t = \sigma(W_{xo}x_t + W_{ho}h_{t-1} + W_{co}c_{t-1}+b_o)\\
            h_t = o_t\text{tanh}(c_t).
        \end{split}
    \end{equation}
    However, as the network implements a FastLSTM layer rather than a conventional LSTM layer, the composite functions are slightly altered \cite{mlpack_fastlstm}.
    \begin{equation}
        \begin{split}
            i_t = \sigma(W_{xi}x_t + W_{hi}h_{t-1} + b_i)\\ 
            f_t = \sigma(W_{xf}x_t + W_{hf}h_{t-1} + b_f)\\ 
            c_t = f_tc_{t-1} + \text{tanh}(W_{xc}x_t + W_{hc}h_{t-1} + b_c)\\ 
            o_t = \sigma(W_{xo}x_t + W_{ho}h_{t-1} + b_o)\\ 
            h_t = \text{tanh}(c_t). 
        \end{split}
    \end{equation}
    \item \textit{Output layer with Mean Squared Error (MSE) performance function}\\
    The output layer evaluates the neural network, in case it is used for training purposes. If the neural network is used to predict a values, the value that is returned reflects the output of the specified output layer. In this case, the MSE function is used to evaluate the network. 
    It does that as follows:
    \begin{equation}
     MSE = \dfrac{1}{m * n} \sum\limits_{i=1}^m \sum\limits_{j=1}^n (x_{ij} - \hat{x}_{ij})^2   
    \end{equation}
    where $x_{ij}$ is the actual value, $\hat{x}_{ij}$ is the predicted value in row $i$ and column $j$, $m$ is the total amount of rows, and $n$ is the total amount of columns.
\end{enumerate}

\noindent The network also uses the Adam optimizer, introduced by \citet{kingma2014adam}, to speed up the training process. The Adam optimizer is computationally efficient, requires little memory, is invariant to diagonal re-scaling of the gradients, and is well suitable for large quantities of data, or problems with many parameters. Therefore, it is a good fit for the solution. 
The resulting formula for updating parameters is:
\begin{equation}
    \begin{split}
        \hat{m}_t = \dfrac{m_t}{1-\beta_1^t},\\
        \hat{v_t} = \dfrac{v_t}{1-\beta_2^t},\\
        a_t = \dfrac{a*\sqrt{1-\beta_2^t}}{1-\beta_1^t},\\
        \theta_{t+1} = \theta_t - \dfrac{a_t * m_t}{\sqrt{\hat{v}_t} + \epsilon}\hat{m}_t,
    \end{split}
\end{equation}
where $\alpha$ is the step size, $m_t$ is the mean at time $t$, and $v_t$ is the uncentered variance of the gradients at time $t$.

\subsection{Trading Strategies and Prediction}

Technical analysis of digital assets is indeed sensitive to the trading strategy, i.e, the methods of buying and selling based on predefined rules to make trading decisions.
There have been longstanding discourses on how useful trading strategies are in practice \cite{knez1996estimating}. Common trading strategies based on time series pattern analysis, such as momentum and contrarian, as well as other technical analysis strategies have had varying degrees of success \cite{conrad1998anatomy, bessembinder1998market, teixeira2010method}. As such, it was decided to add this element to the experimental AI, giving it the capability of applying four different trading strategies based on technical indicators. Due to the vast number of strategies available and conceivable, priority was given to those that are most commonly adopted by existing trading platforms, such as TradingView\footnote{https://www.tradingview.com/} or Cryptowatch\footnote{https://cryptowat.ch/}. Each trading strategy is briefly discussed below.


Every time period (minutely, hourly, daily) the neural network makes a prediction of the next period's asset price. This value is digested by the chosen trading strategy or strategies to determine a decision signal, which is most commonly facilitated by observing the intersection between two or more moving averages. Volatility of the portfolio value can be contained by setting limits on the minimum and maximum amount of portfolio value that can be traded each time period. Altering these limits and changing the trading strategies determines the level of risk management in trading.

\begin{itemize}

\item Rate-of-Change (ROC)\\
The ROC strategy considers the change of the predicted value relative to the last actual value. When the change is below, in between, or above a lower and upper threshold, a decision is made whether to sell, buy or hold.

\item Relative Strength Indicator (RSI)\\
This strategy tracks upward and downward trends in asset value. A sufficiently long series of positive value changes result in an asset being marked as overbought (indicating the holder should sell) and vice versa.

\item Double Exponential Moving Averages (DEMA)\\
The DEMA strategy utilizes two moving averages to determine whether to buy, hold or sell an asset. Crossovers of these averages indicate buy and sell moments as the momentum changes between long-term and short-term asset value.

\item Moving Average Convergence / Divergence (MACD)\\
MACD is a well-known strategy based on the difference between two moving averages with different lengths, called \emph{slow-} and \emph{fast moving average}. This is then compared to the moving average of the difference itself, called the \emph{signal line}. A decision to buy, hold or sell is made whenever the difference crosses the signal line.

\item Random Walk\\
A controversial yet widely regarded investment theory known as the efficient market hypothesis states that all available information is immediately reflected in the value of an asset, and therefore no consistently profitable trading strategies should exist \cite{basu1977investment}. Consequently, a series of random actions should be able to approximate the results of a series of calculated trades.
\end{itemize}

\subsection{Implementation}
RCURRENCY is implemented in C++ with various libraries making up its core. 
First, the historical data is fetched using the API provided by CryptoCompare. Since the API is currently not rate-limited or metered, this step takes only minutes. Live data is fetched in real-time by creating a WebSocket connection with the exchange (Binance) and fetching the candles after they have formed.

After obtaining the historical data, it is preprocessed and fed to the neural network. This process utilizes two well-known libraries: Armadillo\footnote{http://arma.sourceforge.net/}
and TA-Lib\footnote{http://ta-lib.org/}.
%
The Armadillo library provides highly optimized matrix computation functionality desired for regular data set manipulation. TA-Lib provides all necessary functionality for efficiently calculating technical indicators from stock market data. As such, preprocessing and preparation of the training dataset is completed in under a minute.

The recurrent neural network, as described in Section \ref{sec:model}, is implemented using MLPack \cite{mlpack2018}. MLPack is a fast and flexible C++ machine learning library, implementing various types of configurable neural networks and machine learning methods. It utilizes the Armadillo library for linear algebra computations, allowing for relatively low network training times. Training the network on all of the historical data between May 15th, 2013 and January 31st, 2019 takes approximately 10.5 seconds on a single commodity machine (Apple MacBook Pro model A1502, 2.4GHz Intel Core i5-4258U, 8GB RAM). The prediction of a data point as well as an associated trading decision can be obtained in under a millisecond.

\section{Experimental Evaluation} \label{sec:experiments}
Considering the fact that the training data set is time-dependent, Time Series $k$-fold Cross Validation was applied to assess the accuracy of the AI's predictive capabilities and express them numerically in a meaningful manner. The training data set is divided into a training part (80\%) of size $p$ samples and a validation part (20\%) of size $q$ samples, where each sample represents the values of one time unit.

\begin{figure}[t!]
    \centering
    \includegraphics[width=0.41\textwidth]{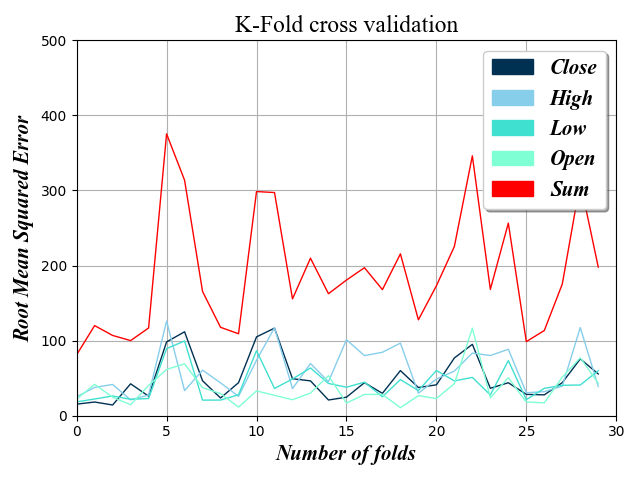}
    \caption{Performance of the network under k-fold cross-validation.}
    \label{fig:k-fold}
\end{figure}

The value $k$ represents how many samples are used in the validation step, or how many time units are predicted in each training-validation cycle. Next, a software environment was developed to facilitate this validation method. The software continuously trained the model on samples $1,\dots,p$, produced a prediction for the next hour $p + 1$, validated the prediction using validation sample $p + 1$, trained the model anew on samples $1,\dots,p + 1$, and so on. Setting the value of $k$ to $1$ means $q$ cycles are performed to complete one series of cross validation. An increasing value of $k$ means a decreasing amount of training-validation cycles. After each cycle, the error is calculated and stored. Finally, an error function is used to measure the performance. A range of values for $k$ was tested in order to obtain the lowest margin of error, as shown in Figure \ref{fig:k-fold} where the Root Mean Squared Error is expressed for the number of chosen folds.

As expected, the AI makes fewer predictive errors for a lower value of $k$, as a larger number of folds reduces the amount of data that is used to train the network before the prediction cycle(s). The figure also shows that the prediction error on all four elements stays well below fifty dollars for lower values of $k$. This effectively allows the system to accurately predict up- and downward trends, as the absolute change between each interval is on average higher than the prediction error. Since the initial plan for the trading engine was to make hourly predictions in the first place, the value of $k$ was fixed to one. A new validation cycle is automatically performed each hour after the JavaScript API pulls new data.

\subsection{Parameter Optimization}
The performance of the neural network is highly dependent on several parameters that fine-tune different aspects of the network. These parameters relate to the topology of the network itself, as well as mathematical functions used to increase the speed of learning and overcome local optima.

To optimize the hyperparameters, the validation technique of the previous section was applied while conducting an exhaustive search for the minimal RMSE by modifying a single parameter at a time. While related works using neural networks often remains silent on how and why specific hyper-parameters have been chosen, in practice they are crucial to the accuracy of the RNN and therefore to the success of any RNN-based agent.

\subsubsection{Edge Weight Initialization}
\begin{figure}[t]
    \centering
    \includegraphics[width=0.41\textwidth]{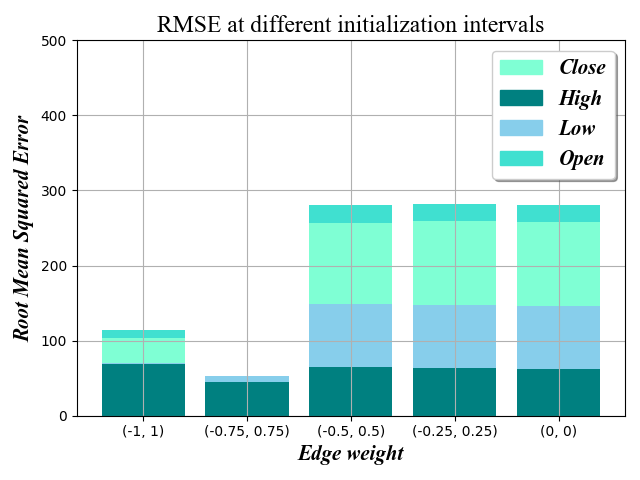}
    \caption{Impact of the different weight initialization values on the RMSE.}
\label{fig:edge-weight}
\end{figure}

The initialization values of these edges will notably influence the training time of the neural network. A more optimal setting of initial weights will lead to a faster and better convergence of the final weight values, and this will reduce the training time significantly \cite{yam2001feedforward}. As Figure~\ref{fig:edge-weight} shows, random numbers in the range of -0.75 and 0.75 proved to have the lowest RMSE across the different folds of the Bitcoin data set.

\begin{figure}[t]
   \subfloat{\label{fig:optimization}
      \includegraphics[width=0.40\textwidth]{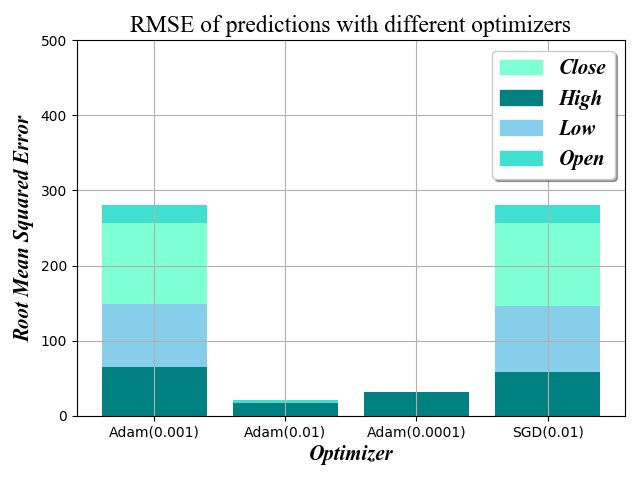}}
    \newline
   \subfloat{\label{fig:rho}
      \includegraphics[width=0.40\textwidth]{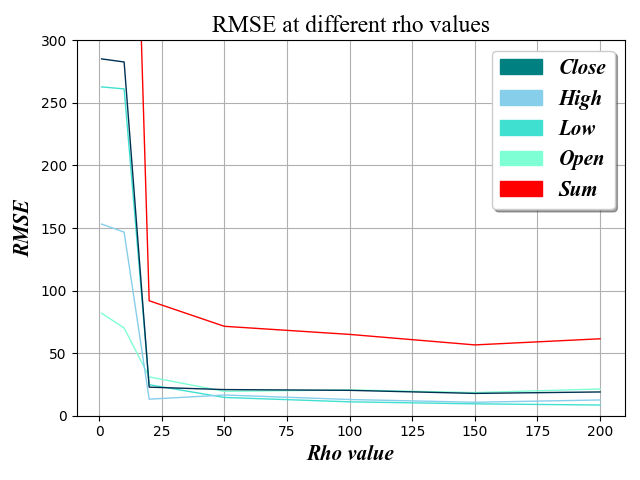}}
    \caption{Impact of parameter optimization on the RMSE.}\label{bs1}
\end{figure}

\subsubsection{Optimization Algorithm}

The role of the optimization algorithm is to adjust the weights on each training iteration of the network, impacting both the speed of convergence as well as the accuracy of the prediction \cite{YAM2000219}. Stochastic Gradient Descent (SGD) methods are the most commonly used optimization algorithms. After testing it with different step sizes against the Adam optimizer \cite{kingma2014adam}, the latter proved to give the best balance of speed and accuracy at a learning rate of 0.01. The results are shown in Figure \ref{fig:optimization}.

\subsubsection{Rho Value (Look-back)}


The rho value sets the length of the back-propagation through time (BPTT) used to train the network. Effectively, it determines how many recent values are used to predict the output at the next step in the time series. As shown in Figure \ref{fig:rho}, the optimal value was found to be around 150.

\subsubsection{Hidden Layer Complexity}

\begin{figure}[t]
    \centering
    \includegraphics[width=0.41\textwidth]{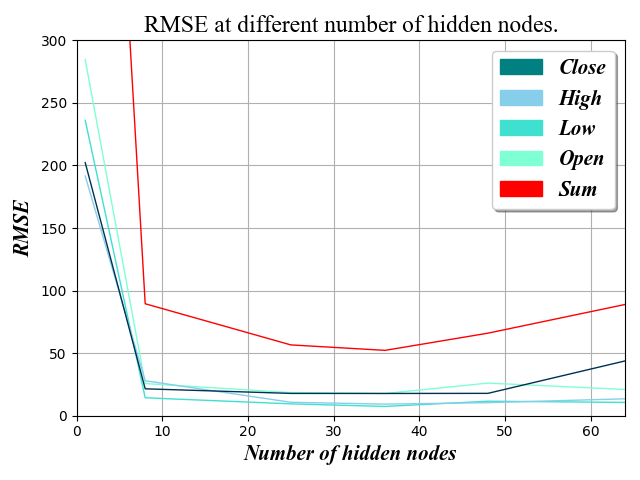}
    \caption{Performance of the different hidden layer sizes, measured against the RMSE.}
\label{fig:hidden-layer}
\end{figure}

The complexity of the hidden layer refers to the number of hidden nodes used in the FastLSTM layer. Increasing the number of nodes brings the potential to learn more complex rules, but at the same time requires more data to train. Furthermore, it increases the risk of over-fitting and thereby losing the ability to generalize, which results in inaccurate results when the network is used with slightly different data sets. Figure~\ref{fig:hidden-layer} shows that the RMSE was the lowest at approximately 36 hidden layer nodes.

\subsection{Evaluating Trading Strategy Results}

This section discusses the results obtained by validating the trading engine and each of its five previously described trading strategies on historical data; a technique popularly known as \emph{backtesting}. The set of historical data points acts as a simulation of live input and updates the portfolio by executing hypothetical ('dummy') trades. The performance of each trading strategy in minimizing losses to the portfolio is dependent on the quality of the prediction used to determine a course of action, as well as the quality of the strategy itself. The recurrent neural network in these experiments utilizes the optimized parameter configuration as mentioned in the previous section to maximize prediction accuracy.

Each experiment concerns trading the BTC/USDT pair, with BTC as the base currency and USDT as the quote currency. The goal is to maximize the portfolio's total monetary value in the quote currency. The total value is calculated by converting the portfolio to quote currency at each period.

\begin{figure} [t] 
    \centering
    \includegraphics[width=0.41\textwidth]{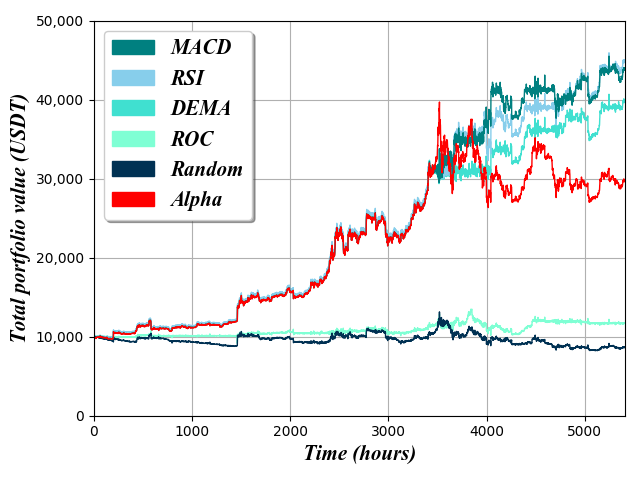}
    \caption{Portfolio growth of 10.000 USDT between January 31th and September 13th, 2019.}
    \label{fig:portfolio-development}
\end{figure}

However, merely achieving a higher monetary value does not imply successful trading if the trader is outperformed by the market itself. If the BTC/USDT value was only ever increasing during a certain time period, then even taking no action whatsoever would have turned a profit. Incidentally, this is known as the \emph{Buy-And-Hold} strategy. With this strategy, a designated asset volume is simply bought and held; no further trading actions are taken. This strategy therefore provides an appropriate baseline for comparison, also known as "trading against Alpha". The results of this comparison and the experiments in general are shown in Figure \ref{fig:portfolio-development}.

\noindent The AI was trained on approximately 90\% of the data set and used to predict the remaining 10\% of the data. Each strategy was tested independently, each receiving an initial portfolio containing 10.000 USDT and subsequently used for trades using the stock market data spanning January 31st to September 13th 2019, approximately 5.400 hours. The Alpha baseline was generated by simply converting the 10.000 USDT into BTC at $t = 0$ and taking no further trading actions. Some limitations have been imposed on the trading agent in order to achieve more stable results. Primarily, a maximum of 25\% of the total portfolio value may be traded at a time. Secondarily, no trades will be made if the expected value increase across the traded volume is smaller than the trading fee (set at 0.1\%).

Figure \ref{fig:portfolio-development} shows that during this part of historical Bitcoin data the RSI strategy managed to achieve the highest portfolio value, with the MACD strategy coming in as a very close second and the DEMA strategy performing only slightly worse. However, all three strategies managed to maintain a stable portfolio value, keeping up with and even outperforming the Alpha baseline. The ROC strategy performs almost as badly as the Random Walk, both barely managing to retain a third of the Alpha baseline value. 

\begin{figure}[t]
    \centering
    \includegraphics[width=0.41\textwidth]{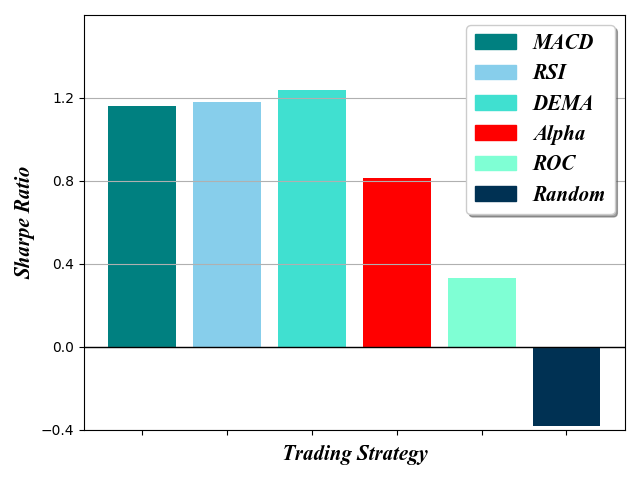}
    \caption{Sharpe Ratios of the six trading strategies between February 13th and September 13th, 2019.}
\label{fig:sharpe-ratios}
\end{figure}

Although the AI is able to outperform the Alpha baseline with certain strategies, the question remains whether the profit merits the risks involved. One way of answering this question is by calculating the Sharpe Ratio \cite{sharpe1966mutual}, which gives an indication of how well an investor is compensated by the returns of an asset compared to a risk-free asset with respect to the risk. The Sharpe Ratio is calculated by subtracting some risk-free asset return rate from the considered asset return rate, divided by the standard-deviation of the considered asset return rate. Figure~\ref{fig:sharpe-ratios} shows the monthly Sharpe Ratios of each of the six trading strategies measured over the final seven (virtual) months of the backtesting period, ranging from February 13th to September 13th, 2019. The return rate $R$ is calculated as the percentage of excess returns relative to the start of each month. As for determining the risk-free rate of return $R_f$, monthly U.S. Treasury Bills (T-Bills) made for an appropriate risk-free asset, as the portfolio is measured in USDT. The T-Bills during the backtesting period had a risk-free rate of roughly 2\%, as indicated on the website of the United States Treasury Department \cite{usgovtreasury2019}. To put these numbers into perspective, a Sharpe Ratio above is 1 is generally considered a good investment, above 2 is considered excellent \cite{goetzmann2002sharpening}. 

The diagram reveals that even though the RSI and MACD strategies attained the highest portfolio values respectively, it is the DEMA strategy that holds the highest Sharpe Ratio. Since the returns ($R$) of the RSI and MACD strategies were higher and the risk-free rate of return ($R_f$) of each month was the same for every strategy, the explanation is that the RSI and MACD strategies displayed a higher volatility (expressed as $\sigma_R$) than the DEMA strategy during this backtesting period. The Random Walk strategy suffered both suffered major losses and a high volatility, thus its Sharpe Ratio fell below negative. Although the ROC strategy suffered similar losses, it managed to keep a positive Sharpe Ratio due to lower volatility levels.
%


\section{Conclusion and Future Work}

With RCURRENCY, we have built a viable AI-based trading agent that is able to operate within the highly volatile cryptocurrency market and even generate profit under real market conditions. RCURRENCY uses a FastLSTM-backed recurrent neural network to predict future stock prices. Network inputs are a combination of direct data (four values representing asset value) and seven values derived from this data (technical indicators), while the four outputs represent the predictions for the four asset values. Performance is measured using the Root Mean Squared Error, which emphasizes the gravity of large prediction errors.
Results obtained from the TSKCV suggest that the network is capable of predicting new asset values with approximately 0.4\% error across the four output channels. The strategies have varying rates of success in maintaining a steady portfolio value throughout the historical dataset but every single one performs better than the randomized strategy. Despite the fact that RCURRENCY is tested for Bitcoin, its design allows the system to support other digital assets as well.

No strategy was able to predict large and sudden drops in asset value correctly. This, however, is a common problem of existing AI-based solutions and future research could shed light on the question whether utilizing a broader set of input data can improve the ability to predict such events effectively. 

Neural networks generally perform better given more training data. One specific input readily available is volume data, which measures what quantities of an asset have been traded in a given period of time. Unfortunately, volume data had to be excluded as it was unknown from which exchanges the volume data originated, a problem that would likely lead to poor predictions. Finding out the origins of the volume data would provide the system with additional training data. 

Despite the fact that RCURRENCY is tested for Bitcoin, the design allows the system to support other digital assets as well. The training data has to include the historical data of the alternative digital asset. Subsequently, the hyper-parameters have to be verified or re-optimized for the newly added data. Another addition, is combining the historical data of multiple assets in different slices per asset, supplying the model with the possibility to utilize the correlations between the slices.



Lastly, it is a common trader's tactic to combine multiple trading strategies to maximize profits. Further simulations could be executed in order to test the performance disparity of the trading engine when combining two, three or even more trading strategies.

\footnotesize{
\bibliographystyle{IEEEtranN}
\bibliography{report.bib}
}

\end{document}